\RequirePackage{fix-cm}
\documentclass[smallextended]{svjour3}       
\smartqed  
\usepackage{bm}
\usepackage{booktabs}
\usepackage{multirow}
\usepackage{color}
\usepackage{lscape}
\usepackage{algorithm}
\usepackage{algorithmic}
\usepackage{graphicx}
\usepackage{amsmath} 
\usepackage{amssymb}  
\begin{document}

\title{Principal Model Analysis Based on Partial Least Squares
}


\author{}
\author{Qiwei Xie         \and
        Liang Tang       \and
        Weifu Li      \and
	        Vijay John \and
            Yong Hu
}

\institute{}

\institute{Qiwei Xie \at
              1. School of Economics and Management, Beijing University of Technology, Beijing, China;\\
           \and
            Liang Tang \at
              2. Research Center for Eco-Environmental Sciences, Chinese Academy of Sciences, China;\\
           \and
           Weifu Li \at
              3. Faculty of Mathematics and Statistics, Hubei University, Wuhan, China;\\
           \and
           Vijay John \at
              4. Toyota Technological Institute, Nagoya, Aichi, Japan;\\
           \and
           Yong Hu \at
              5. Beijing Research Institute of Uranium Geology, Beijing, China.\\
}

\date{Received: date / Accepted: date}

\maketitle

\begin{abstract}
Motivated by the Bagging Partial Least Squares (PLS) and Principal Component Analysis (PCA) algorithms, we propose a Principal Model Analysis (PMA) method in this paper. In the proposed PMA algorithm, the PCA and the PLS are combined. In the method, multiple PLS models are trained on sub-training sets, derived from the original training set based on the random sampling with replacement method. The regression coefficients of all the sub-PLS models are fused in a joint regression coefficient matrix. The final projection direction is then estimated by performing the PCA on the joint regression coefficient matrix. The proposed PMA method is compared with other traditional dimension reduction methods, such as PLS, Bagging PLS, Linear discriminant analysis (LDA) and PLS-LDA. Experimental results on six public datasets show that our proposed method can achieve better classification performance and is usually more stable.
\keywords{Principal model analysis \and partial least squares \and principal component analysis \and dimension reduction \and ensemble learning}
\end{abstract}

\section{Introduction}
\label{intro}
For qualitative analysis, high-dimensional datasets provide enough information,
but in many cases, not all the measured variables are useful for qualitative model.
In addition, traditional statistical methods require the number of variables smaller than the number of samples, otherwise, it will cause the curse of dimensionality \cite{1}. In order to solve these problems, we need to reduce the dimensionality of the dataset before qualitative analysis. Dimension reduction methods such as PCA \cite{2,3,4}, LDA \cite{5} and PLS \cite{6,7} are often used. These methods reduce or eliminate the statistical redundancy and noise between the components of high-dimensional vector data, obtaining a lower-dimensional representation without significant loss of information.

In unsupervised data analysis, PCA is a good tool of dimension reduction, the main idea is to reduce the dimensionality of a dataset in which there are a large number of interrelated variables while retaining as much as possible of the variation present in the dataset \cite{8}. However, PCA can only work in the unsupervised dataset. After adding the sample labels, we need to use supervised methods for analyzing the dataset. LDA is a well-known supervised method for feature extraction and dimension reduction, it achieves maximum discrimination by maximizing the ratio of between-class and within-class distance \cite{9}.
An intrinsic limitation of classical LDA is the so-called small sample size problem \cite{5}, different methods have been proposed to solve this problem \cite{10,11,12}. One of the most successful approaches is subspace LDA, which applies an intermediate dimension reduction stage before LDA. Among all the subspace LDA methods, the PCA plus LDA (PCA-LDA \cite{13}) and PLS plus LDA (PLS-LDA \cite{14}) have received significant attention. Other approaches use the algorithms based on PLS as a dimension reduction.

PLS algorithm has the ability to overcome both the dimensionality and the collinear problems \cite{15,16}, at the same time, and has exhibited excellent performance for solving the problem of small sample size \cite{17}. However, PLS also has some problems, such as how to obtain more useful information, to enhance the robustness of the model, and to more accurately eliminate redundancy and noise. A solution to these problems is ensemble learning which is derived from the field of machine learning \cite{18}, and can be used for both classification and regression problems. In this study, we are more interested in dimensionality reduction and classification. Compared with the single model, ensemble models, including boosting \cite{20,21}, bagging \cite{22} and stacked regression \cite{23,31}, report increased robustness and accuracy \cite{19} and have been successfully applied in the last several years. In order to overcome the over-fitting problem, Zhang et al. used the idea of boosting to combine a set of shrunken PLS models, each with only one PLS component, and called it boosting PLS \cite{20}. On the basis of boosting PLS, some scholars have modified and applied it for spectroscopic quantitative analysis \cite{25,26}.
By using Bagging strategy \cite{27,28}, many training sets are generated from the original dataset, Bagging PLS trains a model from each of those training sets, the final model can be obtained by averaging the coefficient \textbf{B} from each sub-model. From overcoming the disadvantages of MWPLS and iPLS, Xu et al. presented a stack based PLS method using Monte Carlo Cross-validation \cite{29}. Ni et al. have proposed two new stacked PLS which can establish PLS models based on all intervals of a set of spectra to take
advantage of the information from the whole spectrum by incorporating parallel models in a way to emphasize
intervals highly related to the target property \cite{23}.

After the establishment of the PLS sub-models, various ensemble algorithms for the fusion of the final model are available, mainly including average weighting, cross-validation error weighting and minimum square error weighting rule and so on. In this paper, for adopting Bagging model training method, the dataset is divided into a number of sub-training sets.
The PLS models are then employed on these sub-training sets. Subsequently, the coefficients  \textbf{B} of all the PLS sub-models becoming an asymmetric positive semi-definite matrix \textbf{B}\textbf{B}$^{\scriptsize \textmd{T}}$, are fused in a joint matrix. Finally, using the PCA, an eigenvalue decomposition by taking the largest variance model or final projection model is performed. This proposed method is termed as the Principal Model Analysis (PMA). In the subsequent sections, we discussed the relationship between the model parameters (the number of latents, models and remained dimensions) and the classification accuracy. The theory and experiments show that PMA increases the robustness and the generalization ability of the PLS algorithm. Also, PMA can provide a good idea for using the PLS algorithm to semi-supervised dimensionality reduction.

\section{Background}
\label{sec:1}
\subsection{Notation}
\label{sec:2}
Boldface uppercase and lowercase letters are used to denote matrices and vectors, respectively. Lowercase italic letters denote the scalars. The detailed notations are as follows:
\begin{flushleft}
\textbf{X} \qquad $n\times k$ matrix of samples\\
\textbf{y} \qquad  $n\times 1$ vector of sample label\\
\textbf{C} \qquad $k\times k$ covariance matrix of PCA\\
\textbf{w} \qquad $k\times 1$ vector of the PCA loading\\
\textbf{B} \qquad $k\times p$ matrix of PLS regression coefficients\\
\bm{$\beta$} \qquad $k\times 1$ vector of PLS1 regression coefficient\\
$\lambda$ \qquad the eigenvalue of PCA\\
$n$ \qquad number of samples\\
$k$ \qquad number of sample features\\
$m$ \qquad number of components
\end{flushleft}
\subsection{Overview of PLS and Bagging-based PLS}
\label{sec:2}

PLS intends to project the high-dimensional predictor variables into a smaller set of latent variables,
which has a maximal covariance to the responses. Given a training set $(\textbf{X}, \textbf{y})$,
the decomposition of PLS algorithm is as follows \cite{36,37,38,40}:
\begin{displaymath}
\textbf{X}= \sum_{i=1}^{a}\textbf{t}_{i}\textbf{p}_{i}^{\scriptsize \textmd{T}} + \textbf{E},\  \textbf{y} = \sum_{i=1}^{a}\textbf{u}_{i}\textbf{q}_{i}^{\scriptsize \textmd{T}} + \textbf{F},
\end{displaymath}
where $\textbf{t}_{i}$ and $\textbf{u}_{i}$ are score vectors,
and $\textbf{p}_{i}$ and $\textbf{q}_{i}$ are loading vectors of \textbf{X} and \textbf{y}, respectively.
\textbf{E} and \textbf{F} are residuals matrices. The $a$ is the number of feature vectors.
The PLS inner relation between the projected score vectors is:
\begin{displaymath}
\textbf{u}_{i} = \bm{\beta}_{i} \textbf{t}_{i}.
\end{displaymath}
The detailed algorithm procedures of PLS are as follows:
\begin{enumerate}
  \item Initialization: $\textbf{E}_{0}=\textbf{X},\ \textbf{F}_{0}=\textbf{y},\ i=1.$
  \item Computing weight vector: $\textbf{w}_{i}=\textbf{E}_{i-1}^{\scriptsize \textmd{T}}\textbf{F}_{i-1}$, and making $\textbf{w}_{i}$ to be normalization.
  \item Computing the input's score vector: $\textbf{t}_{i}=\textbf{E}_{i-1}\textbf{w}_{i}$ and its loading vector: $\textbf{p}_{i}=\frac{\textbf{E}_{i-1}^{\scriptsize \textmd{T}}\textbf{t}_{i}}{\textbf{t}_{i}^{\scriptsize \textmd{T}}\textbf{t}_{i}}$.
  \item Computing the output's loading vector: $\textbf{q}_{i}=\frac{\textbf{F}_{i-1}^{\scriptsize \textmd{T}}\textbf{t}_{i}}{\textbf{t}_{i}^{\scriptsize \textmd{T}}\textbf{t}_{i}}$, and making $\textbf{q}_{i}$ to be normalization.
  \item Computing the output's score vector: $\textbf{u}_{i}=\textbf{F}_{i-1}\textbf{q}_{i}$.
  \item Computing internal regression coefficient:
  $\bm{\beta}_{i}=\frac{\textbf{u}_{i}^{\scriptsize \textmd{T}}\textbf{t}_{i}}{\textbf{t}_{i}^{\scriptsize \textmd{T}}\textbf{t}_{i}}$.
  \item Computing residuals matrices: $\textbf{E}_{i}=\textbf{E}_{i-1}-\textbf{t}_{i}\textbf{p}_{i}^{\scriptsize \textmd{T}}$ and $\textbf{F}_{i}=\textbf{F}_{i-1}-\bm{\beta}_{i}\textbf{t}_{i}\textbf{q}_{i}^{\scriptsize \textmd{T}}$.
  \item Updating $i$ to $i+1$, then go back the step 2 until the expected number of latent variables is achieved.
\end{enumerate}

The ensemble learning method aims to improve the accuracy and robustness of traditional algorithms by combining the results of multiple sub-models. Bagging is a simple ensemble learning strategy and is widely used for the classification and regression problems, such as bagging SVM and bagging PLS.

The general PLS method usually shows bad or unstable results on the data with a very large number of collinear x-variables or the data with very limited training samples. By using the bagging strategy, the bagging PLS model could reduce the variance of the original unstable model without increasing the bias. Therefore, bagging PLS usually can achieve much more accurate and stable results than traditional PLS method.

Bagging-based PLS first generates several sub-training sets from the original training set based on the random sampling with replacement method,
and then trains a PLS model on each sub-training set separately, finally averages the regression coefficients of all sub-PLS models
and uses the averaged regression coefficient for the model prediction.
In detail, we suppose that $N$ sub-training sets are generated by random sampling with replacement,
and the PLS regression coefficient vector corresponding to each sub-training set is $\bm{\beta}_{i} (i=1,\cdots, N)$.
The final regression coefficient of bagging-based PLS can be formulated as:
\begin{equation}
\bm{\beta} = \frac{1}{N} \sum_{i=1}^{N} \beta_{i}
\end{equation}

\subsection{Overview of PCA}
\label{sec:2}
Principal component analysis (PCA) uses an orthogonal transformation to convert a number of possibly correlated variables into a smaller number of uncorrelated variables called principal components while trying to preserve the data variance. Given a data matrix  \textbf{X}, computing the covariance matrix \textbf{C}, then the projection directions of PCA can be solved by:
\begin{equation}
\hat{\textbf{w}} = \arg \max_{\|\textbf{w}\|=1} \textbf{w}^{\scriptsize \textmd{T}}\textbf{C}\textbf{w}.
\end{equation}
The above problem can be easily solved by the eigendecomposition methods, such as the singular-value decomposition (SVD) algorithm.
The detailed algorithmic processes of PCA is as follows \cite{41,42,43,44}:
\begin{enumerate}
  \item Data standardization:
\begin{displaymath}
x_{ij}^{*}=\frac{x_{ij}-\bar{x}_{j}}{s_{ij}},\ s_{ij}=\sqrt{\frac{\sum_{j=1}^{m}(x_{ij}-\bar{x}_{j})^{2}}{s_{ij}}},
\end{displaymath}
where $\textbf{X}=(x_{ij})_{n\times p}$ is the data matrix with $n$ samples and $p$ variables.
  \item Computing the covariance matrix: $\textbf{C}=\textbf{X}^{\scriptsize \textmd{T}}\textbf{X}$.
  \item Eigendecomposition:  $\textbf{C}\textbf{u}=\lambda \textbf{u}$.
  \item Denote the first $m$ eigenvalues as $\lambda_{1} \geq \lambda_{2} \geq \ldots \geq \lambda_{m}$, their corresponding eigenvectors,
$\textbf{u}_{1}, \textbf{u}_{2}, \cdots, \textbf{u}_{m}$, are principal components. The number of principal components $m$ can be decided by the cumulative contribution rate of the principal components, i.e., choosing $m$ such that
$$\frac{\sum_{i=1}^{m}\lambda_{i}}{\sum_{j=1}^{p}\lambda_{j}} \geq 0.95.$$
\end{enumerate}
\section{Principal Model Analysis}
\subsection{Theory and Algorithm}
Combing the bagging strategy and PLS, we propose a principal model analysis (PMA) method in this section.
The proposed PMA contains two steps. The first step is also to  generate $\widehat{N}$ sub-training sets  from the original training set
with replacement method and
the corresponding PLS regression coefficient vector of each sub-training set is denoted by $\bm{\beta}_{i}, i=1,\cdots,\widehat{N}$.
Different from the bagging PLS method which just simply averages the PLS sub-models,
the second step  adopted here is to use the PLS sub-models as the input of PCA algorithm to generate the final PMA model.
It is mainly because PCA can effectively find the ``major" elements,  remove the noise,
and reveal the essential structure hidden behind the complex data.

The original PCA algorithm is performed by decomposing the covariance matrix  \textbf{C},
which is a symmetric and positive semi-definite matrix. However, the whole regression coefficient matrix
$\textbf{B}=[\bm{\beta}_{1},\ldots,\bm{\beta}_{n}]$ in the PMA algorithm is not a symmetric and positive semi-definite matrix.
So, we need to make the regression coefficient matrix \textbf{B} to be a symmetric positive semi-definite matrix. We replace the \textbf{B}
by \textbf{BB}$^{\scriptsize \textmd{T}}$ for the eigenvalue decomposition,
and get the most representative models which called principal model
as the final PMA model. The optimization of PMA algorithm can be expressed as:
\begin{equation}\label{1_3}
\left\{ \begin{array}{ll}
\textbf{B} = [\bm{\beta}_{1},\ldots,\bm{\beta}_{n}][\bm{\beta}_{1},\ldots,\bm{\beta}_{n}]^{\scriptsize \textmd{T}}\\
\hat{\textbf{w}} =  \arg \max_{\|\textbf{w}\|=1} \textbf{w}^{\scriptsize \textmd{T}}\textbf{B}\textbf{w}
\end{array} \right.
\end{equation}

The above problem can be easily solved by the singular-value decomposition (SVD) algorithm:
\begin{enumerate}
  \item Eigendecomposition: $\textbf{B}\textbf{u}=\lambda \textbf{u}$.
  \item Denote the first $m$ eigenvalues as  $\lambda_{1} \geq \lambda_{2} \geq \ldots \geq \lambda_{m}$,
  their corresponding eigenvectors, $\textbf{u}_{1}, \textbf{u}_{2}, \cdots, \textbf{u}_{m}$, are principal components.
  The number of principal components $m$ can be decided by the cumulative contribution rate of the principal components, i.e., choosing $m$ such that
  $$\frac{\sum_{i=1}^{m}\lambda_{i}}{\sum_{j=1}^{p}\lambda_{j}} \geq 0.95.$$
\end{enumerate}

The detailed processes of PMA method are shown as follows.\\\newline
\noindent\rule[0.25\baselineskip]{\textwidth}{1pt}
\textbf{Algorithm PMA}\newline
\textbf{Input:} Training set and corresponding label vector, the number of PLS latent variables,
the number of sub-models, the number of principal models (dim).\newline
\textbf{Output:} The projection direction of PMA.\newline
\noindent\rule[0.25\baselineskip]{\textwidth}{1pt}
1. Preprocessing the training set $\mathcal{T}$.\newline
2. Dividing $\mathcal{T}$ using random sampling with replacement and generating PLS sub-models $\bm{\beta}_{i}$.\newline
3. Denote $\textbf{B}=[\bm{\beta}_{1},\cdots,\bm{\beta}_{n}][\bm{\beta}_{1},\cdots,\bm{\beta}_{n}]^{\scriptsize \textmd{T}}$, and doing the eigenvalue decomposition in (\ref{1_3}), sorting the eigenvalues in descending order and rearranging their corresponding eigenvectors.\newline
4. Denote the rearranged eigenvector matrix as \textbf{W}, outputting the final PMA model $B_{PMA}=\textbf{W}(:,1:\textmd{dim})$.\newline
\noindent\rule[0.25\baselineskip]{\textwidth}{1pt}
\subsection{Determination of the number of Latent variables}
The number of latent variables is an important parameter in the PLS model.
There are many approaches to determine the number of latent variables, such as genetic algorithm,
F-test and cross-validation methods. Cross-validation methods include K-fold cross-validation (k-CV),
leave-one-out cross-validation (LOOCV), Monte Carlo cross-validation (MCCV) and so on.
In this paper, we use 10-fold cross-validation method to determine the number of latent variables.
\subsection{The sub-models selective rule}
For ensemble strategy, usually sub-models who performed better or part of the performance can include more diversity \cite{27}.
So Zhou et al. suggested that it may be a better choice for using part of sub-models instead of all of the sub-models \cite{32}.
Herein, original training set is arbitrarily divided into tress parts: calibration sets, validation sets and prediction sets \cite{45}.
We establish 100 PLS sub-models in the validation sets by sub-sampling and re-weighting the existed calibration samples respectively. The proposed method directly constructs diverse models with virtual samples which are produced by original calibration samples, and this can increase the amount of ensemble diversity when the calibration samples are not enough \cite{45}. Using these 100 sub-models on the validation set,
we get 100 different classification accuracies. Then follow the classification accuracy in descending order,
take the sub-models with largest classification accuracy participate final ensemble.
\subsection{Determination of Dimensions}
PCA does an EIG or SVD on a matrix and then generates an eigenvalue matrix. To select the principal components we have to take only the first few eigenvalues. How do we decide on the number of eigenvalues that we should take from the eigenvalue matrix? Usually we adopt accumulative contribution rate automatically retain useful eigenvalues.

Using PMA to reduce dimension is to obtain the scores by projecting the new samples to the direction of the principal models, so the number of the final dimensions is equal to the number of selected principal models. In the experiment, if the parameter of fixed dimensions, which is one of inputs, is greater than or equal to 1, we will use fixed dimensions to select the principal models. Otherwise, if the fixed dimensions greater than 0 and less than 1, we use cumulative contribution rate to obtain the principal models.

From the selection of sub-models can be inferred, the number of final principal models does not need much. Because the classification ability of the selected sub-models are almost the same, so only one principal model almost retains all sub-models classification ability. Therefore, in practical applications, we only take the eigenvector with the largest eigenvalue as the principal model.

\section{Experimental results}
\subsection{Data Sets}
In order to evaluate the performance of the proposed PMA method, we compare it with the PLS, LDA, PLS-LDA
and Bagging PLS methods on three types of data sets:
\begin{enumerate}
  \item Four UCI datasets, i.e., Breast data, Spambase data,
Gas data, Musk data (Version 1) (obtained from http://archive.ics.uci.edu/ml/datasets.html);
  \item Small data and Imbalanced data;
  \item Raman spectral data (Raman).
\end{enumerate}
The details of these datasets are shown in Table \ref{tab:1}.
Before using the datasets, we remove the non-numerical and missing inputs data and convert the class label to a numeric type.
\begin{table}[!htb]
\caption{Data Sets}
\label{tab:1}       
\begin{tabular}{lccccccccc}
\hline\noalign{\smallskip}
Data Set & Number of Examples & Number of Attributes & Class label & Year  \\
\noalign{\smallskip}\hline\noalign{\smallskip}
Breast & 569 & 30 & 1 and 2 & 1995 \\
Spambase & 4601 & 57 & 0 and 1 & 1999 \\
Gas & 4782 & 128 & 5 and 6 & 2012 \\
Musk(Version 1) & 168 & 476 & 0 and 1 & 1994 \\
small & 300 & 476 & 0 and 1 & 1994 \\
imbalanced & 7074 & 476 & 0 and 1 & 1994 \\
Raman & 925 & 101 & 0 and 4 & N/A \\
\noalign{\smallskip}\hline
\end{tabular}
\end{table}

The data sets ``small" and ``imbalanced" are randomly sampled from the data set ``Musk (Version 1)".
The data set ``small" is a typical data set with high dimensionality and small samples, where the number of positive and negative samples are the same.
The data set ``imbalanced"  is an imbalanced data set, where the ratio of positive and negative samples is 6:1.

Spectral data set ``Raman" is obtained by a standard Raman spectroscope (HR LabRam invers, Jobin-Yvon-Horiba).
The excitation wavelength of the used laser (Nd: YAG, frequency doubled) is centered at 532 nm. There are 2545 spectra for 20 different strains available \cite{34}. Herein we select two classes (B. subtilis DSM 10 and M. luteus DSM 348) and use the spectra in the region 1100-1200 in calculations \cite{35}.

\subsection{Calculation}
Five dimension reduction methods, i.e., PLS, LDA, PLS-LDA, Bagging PLS and PMA, are compared in our experiments. For Bagging PLS algorithm, fifteen models are generated by the random sampling with replacement method and the final model is obtained by averaging these fifteen sub-models. For the PMA method, 100 sub-models are generated from the validation set, and the best fifteen sub-models with higher accuracies are chosen to perform model fusion. Except for the LDA, the number of latent variables in the PLS, PLS-LDA, Bagging PLS and PMA are determined by the 10-fold cross-validation. In the experiment, the dimensionality of the original data is reduced to 1. For fair comparison, the linear Naive Bayes classifier is used to evaluate the results of the above different dimension reduction methods.

For each data set, we randomly choose 49$\%$, 30$\%$ and 21$\%$ samples from the total samples to form the training set, test set, and validation set. The experiments are randomly run 20 times, and the averaged results are recorded.


\section{Results and Discussion}
\subsection{Classification performance of different algorithms}
This section mainly investigates the classification performance of various algorithms.
We report the results on both the training and testing datasets.
The classification accuracies accuracies are reported in Table \ref{tab:2} and Table \ref{tab:3}, respectively.

\begin{table}[!htb]
\caption{Train Classification Accuracy of Different Comparative Algorithms}
\label{tab:2}       
\begin{tabular}{lccccccccc}
\hline\noalign{\smallskip}
Data Set & PLS & LDA & PLS-LDA & Bagging PLS & PMA  \\
\noalign{\smallskip}\hline\noalign{\smallskip}
    Breast & 0.9183  & 0.6750  & 0.8910  & 0.9307  & \textbf{0.9632 } \\
    Spambase & 0.8043  & 0.7763  & 0.6816  & 0.8543  & \textbf{0.9070 } \\
    Gas   & 0.9704  & 0.9715  & 0.7920  & 0.9703  & \textbf{0.9740 } \\
    Musk (Version 1) & 0.9119  & 0.7176  & 0.9039  & 0.9126  & \textbf{0.9176 } \\
    small & 0.9249  & 0.9037  & 0.9299  & 0.9481  & \textbf{0.9858 } \\
    imbalanced & 0.9726  & \textbf{1.0000 } & 0.9619  & 0.9769  & 0.9899  \\
    Raman & 0.9515  & 0.7654  & 0.8508  & 0.9538  & \textbf{0.9579 } \\
\noalign{\smallskip}\hline
\end{tabular}
\newline
The bold value means the maximum accuracy among different methods.
\end{table}
\begin{table}[!htb]
\caption{Test Classification Accuracy of Different Comparative Algorithms}
\label{tab:3}       
\begin{tabular}{lccccccccc}
\hline\noalign{\smallskip}
Data Set & PLS & LDA & PLS-LDA & Bagging PLS & PMA  \\
\noalign{\smallskip}\hline\noalign{\smallskip}
Breast & 0.9143  & 0.6428  & 0.8891  & 0.9265  & \textbf{0.9545 } \\
    Spambase & 0.8025  & 0.7607  & 0.6820  & 0.8522  & \textbf{0.9034 } \\
    Gas   & 0.9694  & 0.9623  & 0.7920  & 0.9694  & \textbf{0.9730 } \\
    Musk (Version 1) & 0.9052  & 0.7003  & 0.8980  & 0.9059  & \textbf{0.9108 } \\
    small & 0.7108  & 0.6136  & 0.7128  & 0.7215  & \textbf{0.7220 } \\
    imbalanced & 0.9003  & 0.6492  & 0.8821  & 0.9091  & \textbf{0.9097 } \\
    Raman & 0.9345  & 0.6589  & 0.8400  & 0.9367  & \textbf{0.9419 } \\
\noalign{\smallskip}\hline
\end{tabular}
\newline
The bold value means the maximum accuracy among different methods.
\end{table}

The small-sample-size problem is often encountered in the field of pattern recognition.
It may lead to the singularity of the within-class scatter matrix in the LDA. So, for the data sets ``small" and ``imbalanced",
the LDA algorithm shows bad results. PLS shows good overall classification performance.
PLS-LDA algorithm firstly removes redundancy and noise in the data set by the PLS method,
then performs the LDA algorithm on the PLS dimension reduction features.
PLS-LDA shows better results than LDA except for the data sets ``Spambase" and ``Gas".
But PLS-LDA still seems to show over-fitting phenomenon in the data sets ``small" and ``imbalanced".
Because the PLS dimension reduction process may lose some information, the results of PLS-LDA are worse than PLS.
The Bagging PLS achieves better results than PLS in the data sets ``Breast", ``Spambase", ``Raman" and ``Muskv (Version 1)".
Although many sub-models in Bagging PLS provide better performance than PLS, the improvement of Bagging PLS over PLS is
not significant because of the average strategy. As observed in the Tables \ref{tab:2} and \ref{tab:3},
all algorithms appear over-fitting phenomenon on  the  data set ``small".
Notwithstanding, the proposed PMA algorithm achieves the best results in either training and testing set except
for the  data set ``imbalanced". The superiorities are much more obvious on the data sets ``Breast" and ``Spambase" .

\begin{figure}
  \includegraphics[scale=0.5]{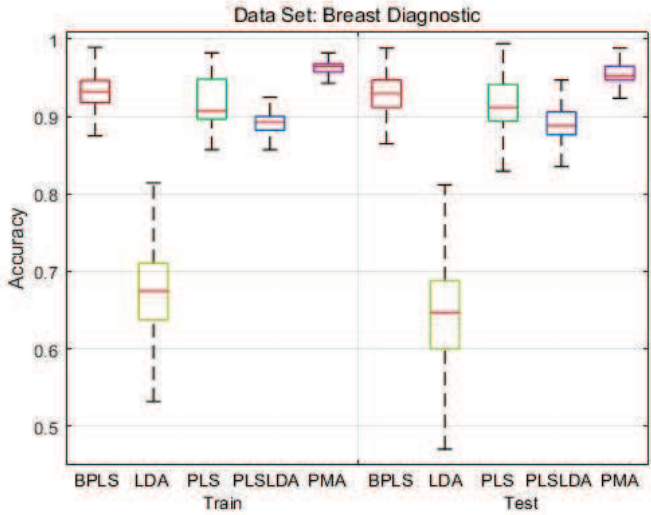}
\caption{Classification accuracy box of Breast Diagnostic data}
\label{fig:f1}       
\end{figure}

\begin{figure}
  \includegraphics[scale=0.5]{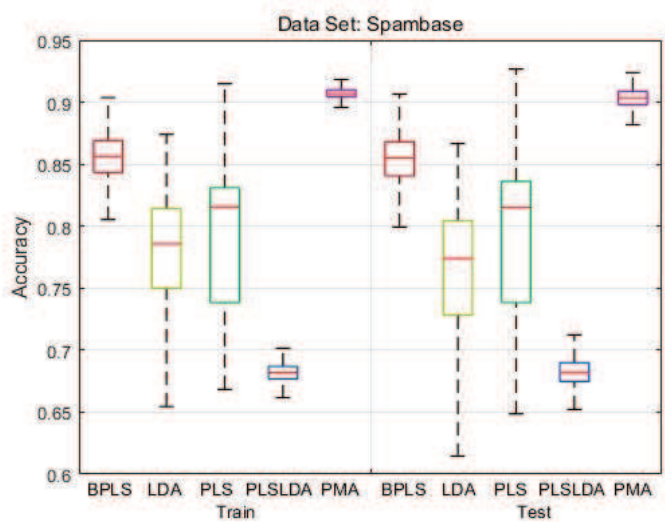}
\caption{Classification accuracy box of Spambase data}
\label{fig:f2}       
\end{figure}

\begin{figure}
  \includegraphics[scale=0.5]{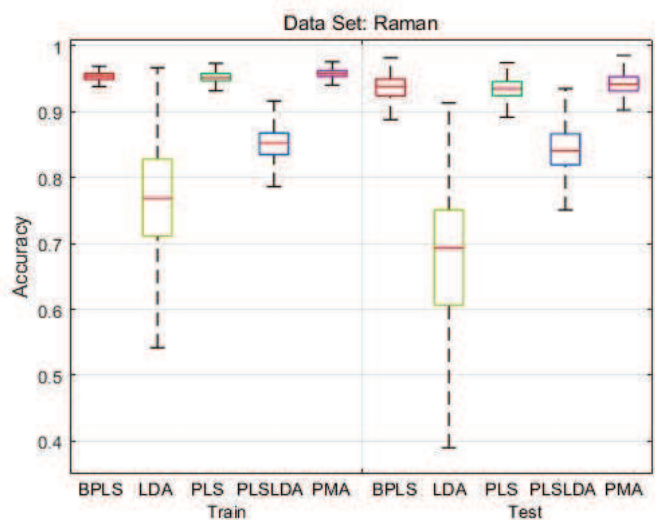}
\caption{Classification accuracy box of Raman data}
\label{fig:f3}       
\end{figure}

\begin{figure}
  \includegraphics[scale=0.5]{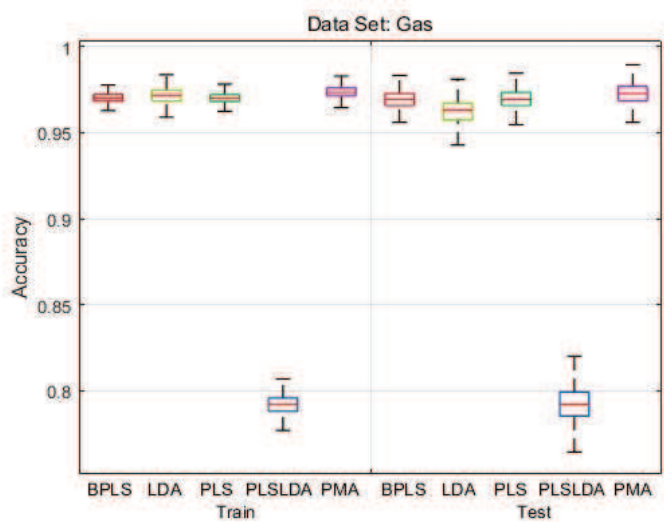}
\caption{Classification accuracy box of Gas data}
\label{fig:f4}       
\end{figure}

\begin{figure}
  \includegraphics[scale=0.5]{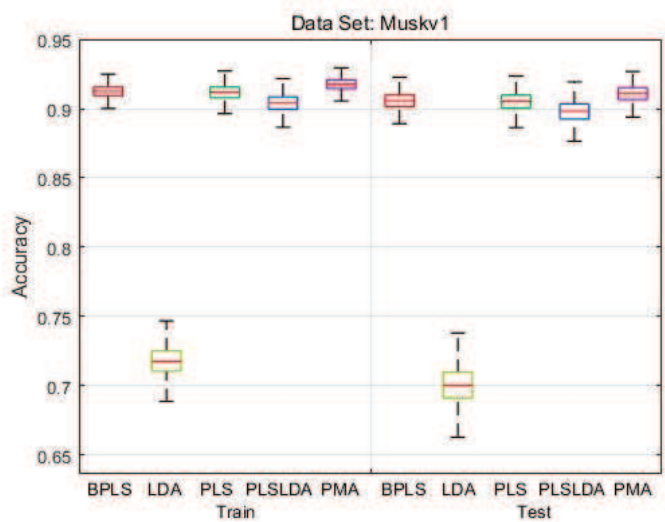}
\caption{Classification accuracy box of Muskv1 data}
\label{fig:f5}       
\end{figure}

\begin{figure}
  \includegraphics[scale=0.5]{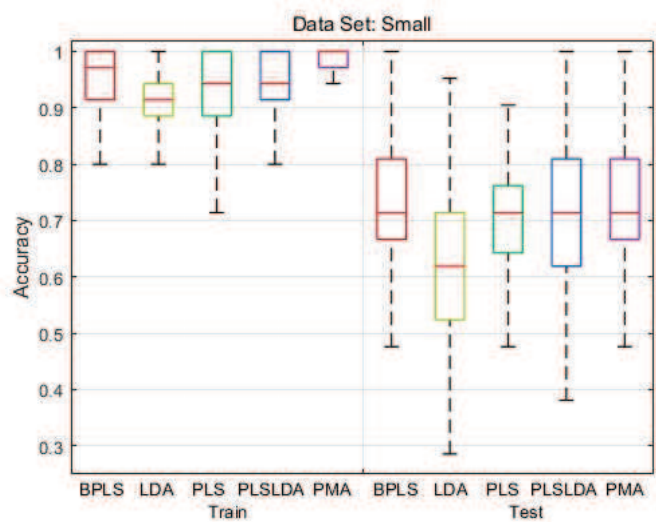}
\caption{Classification accuracy box of Small data}
\label{fig:f6}       
\end{figure}

\begin{figure}
  \includegraphics[scale=0.5]{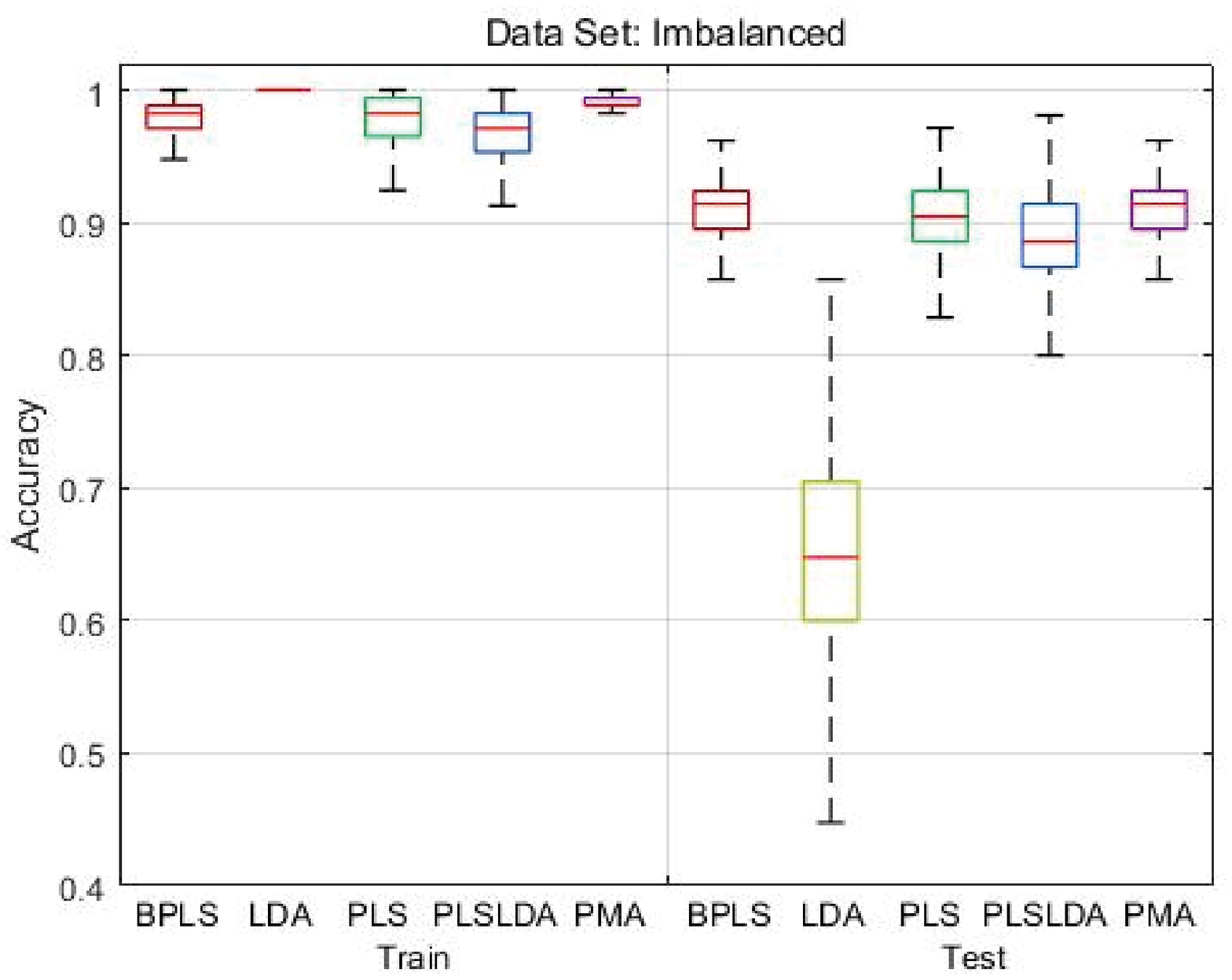}
\caption{Classification accuracy box of Imbalance data}
\label{fig:f7}       
\end{figure}

The above figures are the box diagrams of the accuracy of different algorithms.
For the data sets ``Breast", ``Spambase", ``Raman" and ``Muskv (Version 1)", the results of LDA algorithm are obviously much worse than other methods. The results of PLS are also unstable on the data set ``Spambase". The results of PLS-LDA are worse than others on the data sets ``Spambase" and ``Gas". In the data set ``small", all algorithms show over-fitting phenomenon.
Except for the data set ``small", PMA algorithm gets more stable results than other methods.

\subsection{Investigation on the number of sub-models}
\begin{figure}[!htbp]
  \includegraphics{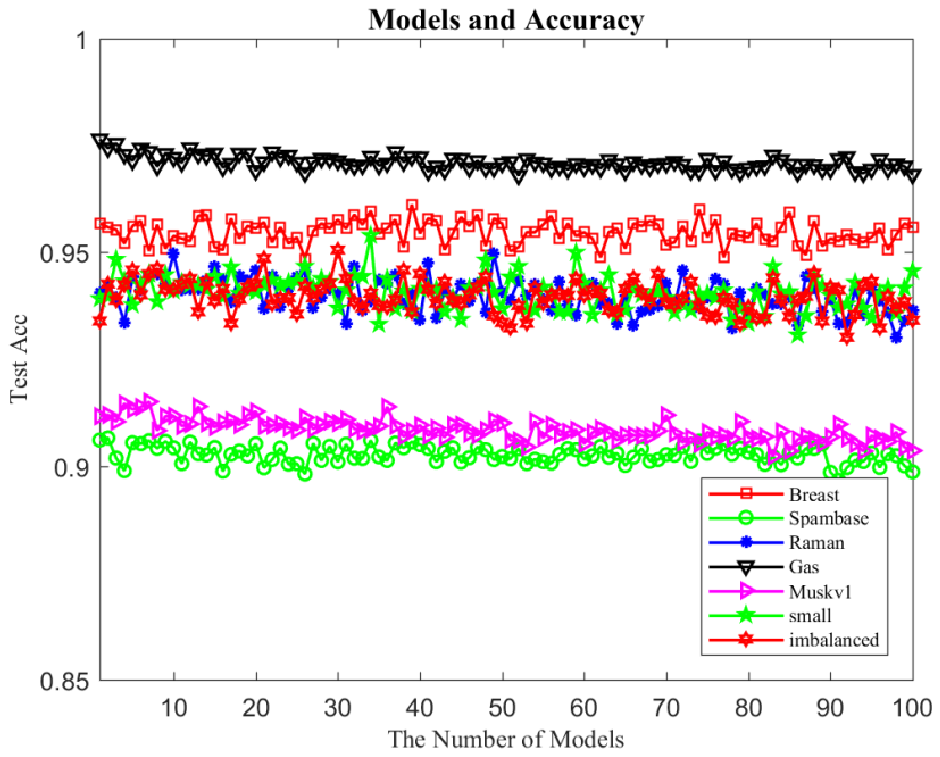}
\caption{Classification accuracy vs.number of sub-models}
\label{fig1}       
\end{figure}

From the Figure \ref{fig1}, we can see that the number of sub-models is less sensitive to the PMA model.
In general, the classification accuracies on each data set decreases with the increase of the number of sub-models.
It demonstrates that not all of the sub-models are valid.
Meanwhile, it is likely to improve the classification performance by choosing some good sub-models.
For the data sets ``Breast" and ``Raman", the number of sub-models greatly affects the classification results.
The number of sub-models can be empirically determined by the cross-validation method.

\subsection{Impact of PMA dimensions}
\begin{figure}[!htbp]
  \includegraphics{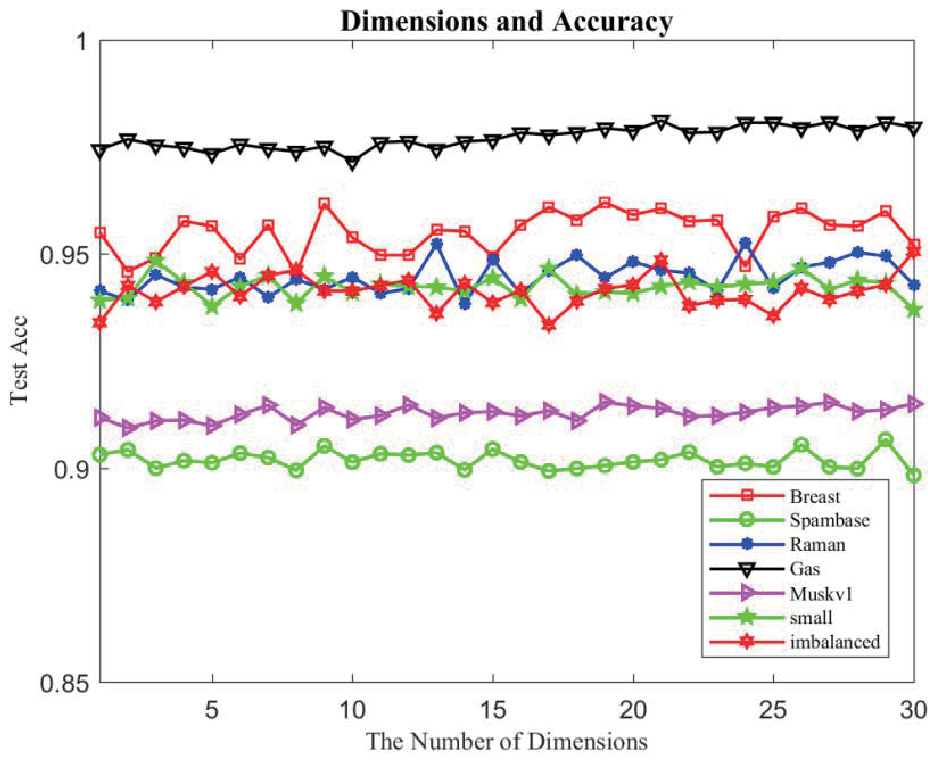}
\caption{The impact of PMA demensions}
\label{fig2}       
\end{figure}
To investigate the effect of PMA dimensionalities, we show the classification results on different dimensionalities ranging from 1 to 30.
As can be seen from the Figure \ref{fig2}, the classification accuracy on all data sets does not improve with the increase of dimensionality.
A possible explanation could be that the first principal component already contains the majority information of the entire data.
The results on the data sets ``Gas" and ``Muskv1" are relatively stable.
\subsection{Discussions of the proposed method}
The proposed PMA algorithm extends the original Bagging PLS for qualitative analysis. The results on the six data sets show that PMA algorithm can improve the classification accuracy to a certain extent. Model ensemble has many advantages, such as enhancing the robustness. However, the number of sub-models and the number of dimensionalities must be carefully chosen.

\section{Conclusions}
In this paper, we have proposed a PMA method for classification. By means of ensemble strategy, the proposed PMA method fuses the results of PLS sub-models and finds the principal model by performing PCA on the joint coefficient matrix of all sub-models. Experimental results demonstrate that the proposed PMA method can achieve better classification performance than original PLS and Bagging PLS. Our future work will focus on finding more comprehensive evaluation criteria for the selection of sub-models. In addition, we will perform PMA on semi-supervised problems by adding a large number of unsupervised data.


\bibliographystyle{spmpsci}      
\bibliography{mybibfile}   

\begin{thebibliography}{10}
\providecommand{\url}[1]{{#1}}
\providecommand{\urlprefix}{URL }
\expandafter\ifx\csname urlstyle\endcsname\relax
  \providecommand{\doi}[1]{DOI~\discretionary{}{}{}#1}\else
  \providecommand{\doi}{DOI~\discretionary{}{}{}\begingroup
  \urlstyle{rm}\Url}\fi

\bibitem{22}
Afara, I., Singh, S., Oloyede, A.: Application of near infrared (nir)
  spectroscopy for determining the thickness of articular cartilage.
\newblock Medical Engineering and Physics \textbf{35}(1), 88--95 (2013)

\bibitem{6}
Barker, M., Rayens, W.: Partial least squares for discrimination.
\newblock Journal of Chemometrics \textbf{30}(3), 446--452 (2012)

\bibitem{1}
Bellman, R.: Adaptive Control Processes: A Guided Tour.
\newblock The University Press (1961)

\bibitem{31}
Bi, Y., Xie, Q., Peng, S., Tang, L., Hu, Y., Tan, J., Zhao, Y., Li, C.: Dual
  stacked partial least squares for analysis of near-infrared spectra.
\newblock Analytica Chimica Acta \textbf{792}(16), 19--27 (2013)

\bibitem{38}
Bian, X., Li, S., Shao, X., Liu, P.: Variable space boosting partial least
  squares for multivariate calibration of near-infrared spectroscopy.
\newblock Chemometrics and Intelligent Laboratory Systems \textbf{158} (2016)

\bibitem{15}
Boulesteix, A.L.: Pls dimension reduction for classification with microarray
  data.
\newblock Statistical Applications in Genetics and Molecular Biology
  \textbf{3}(1), 392 (2004)

\bibitem{27}
Breiman, L.: Bagging predictors.
\newblock Machine Learning \textbf{24}(2), 123--140 (1996)

\bibitem{10}
Chen, L.F., Liao, H.Y.M., Ko, M.T., Lin, J.C., Yu, G.J.: A new lda-based face
  recognition system which can solve the small sample size problem.
\newblock Pattern Recognition \textbf{33}(10), 1713--1726 (2000)

\bibitem{44}
Chiang, K.Y., Hsieh, C.J., Dhillon, I.S.: Robust principal component analysis
  with side information.
\newblock In: International Conference on Machine Learning, pp. 2291--2299
  (2016)

\bibitem{28}
Efron, Bradley: An introduction to the bootstrap.
\newblock Chapman and Hall (1995)

\bibitem{35}
Ferrari, A.C., Robertson, J.: Interpretation of raman spectra of disordered and
  amorphous carbon.
\newblock Physical Review B \textbf{61}(20), 14,095--14,107 (2000)

\bibitem{40}
Folch-Fortuny, A., Arteaga, F., Ferrer, A.: Pls model building with missing
  data: New algorithms and a comparative study.
\newblock Journal of Chemometrics \textbf{31}(1-2) (2017)

\bibitem{3}
Ginkel, J.R.V., Kroonenberg, P.M.: Using generalized procrustes analysis for
  multiple imputation in principal component analysis.
\newblock Journal of Classification \textbf{31}(2), 242--269 (2014)

\bibitem{17}
Goodhue, D.L., Lewis, W., Thompson, R.: Does pls have advantages for small
  sample size or non-normal data?
\newblock Mis Quarterly \textbf{36}(3), 981--1001 (2012)

\bibitem{45}
Hu, Y., Peng, S., Peng, J., Wei, J.: An improved ensemble partial least squares
  for analysis of near-infrared spectra.
\newblock Talanta \textbf{94}, 301--307 (2012)

\bibitem{11}
Huang, R., Liu, Q., Lu, H., Ma, S.: Solving the small sample size problem of
  lda \textbf{3}, 29--32 (2002)

\bibitem{42}
Jolliffe, I.T., Cadima, J.: Principal component analysis: a review and recent
  developments.
\newblock Philosophical Transactions \textbf{374}(2065), 20150,202 (2016)

\bibitem{8}
Kambhatla, N., Leen, T.: Dimension reduction by local principal component
  analysis.
\newblock Neural Computation \textbf{9}(7), 1493--1516 (1997)

\bibitem{7}
Liu, Y., Rayens, W.: Pls and dimension reduction for classification.
\newblock Computational Statistics \textbf{22}(2), 189--208 (2007)

\bibitem{36}
Long, C., Guizeng, W.: Soft sensing based on pls with iterated bagging method.
\newblock Journal of Tsinghua University \textbf{48}, 86--90 (2008)

\bibitem{18}
Maclin, R., Opitz, D.: Popular ensemble methods: An empirical study.
\newblock Journal of Artificial Intelligence Research \textbf{11}, 169--198
  (2011)

\bibitem{14}
Marigheto, N.A., Kemsley, E.K., Defernez, M., Wilson, R.H.: A comparison of
  mid-infrared and raman spectroscopies for the authentication of edible oils.
\newblock Journal of the American Oil Chemists' Society \textbf{75}(8),
  987--992 (1998)

\bibitem{4}
Martinez, A.M., Kak, A.C.: Pca versus lda.
\newblock IEEE Transactions on Pattern Analysis and Machine Intelligence
  \textbf{23}(2), 228--233 (2001)

\bibitem{19}
Mendes-Moreira~J Soares~C, J.A.M.: Ensemble approaches for regression: A
  survey.
\newblock ACM Computing Surveys \textbf{45}(1), 10 (2011)

\bibitem{5}
Montanari, A.: Linear discriminant analysis and transvariation.
\newblock Journal of Classification \textbf{21}(1), 71--88 (2004)

\bibitem{23}
Ni, W., Brown, S.D., Man, R.: Stacked partial least squares regression analysis
  for spectral calibration and prediction.
\newblock Journal of Chemometrics \textbf{23}(10), 505--517 (2010)

\bibitem{34}
Peschke, K.D., Haasdonk, B., Ronneberger, O., Burkhardt, H., Harz, M.: Using
  transformation knowledge for the classification of raman spectra of
  biological samples.
\newblock In: Proceedings of the 4th Iasted International Conference on
  Biomedical Engineering, pp. 288--293 (2006)

\bibitem{26}
Qin, X., Gao, F., Chen, G.: Wastewater quality monitoring system using sensor
  fusion and machine learning techniques.
\newblock Water Research \textbf{46}(4), 1133--1144 (2012)

\bibitem{37}
Ren, D., Qu, F., Lv, K., Zhang, Z., Xu, H., Wang, X.: A gradient descent
  boosting spectrum modeling method based on back interval partial least
  squares.
\newblock Neurocomputing \textbf{171}(C), 1038--1046 (2012)

\bibitem{21}
Shao, X., Bian, X., Cai, W.: An improved boosting partial least squares method
  for near-infrared spectroscopic quantitative analysis.
\newblock Analytica Chimica Acta \textbf{666}(1-2), 32--37 (2010)

\bibitem{41}
Shao-Hong, G.U., Wang, Y.S., Wang, G.X.: Application of principal component
  analysis model in data processing.
\newblock Journal of Surveying and mapping \textbf{24}(5), 387--390 (2007)

\bibitem{25}
Tan, Chao, Wang, Jinyue, Wu, Tong, Qin, Xin, Li, Menglong: Determination of
  nicotine in tobacco samples by near-infrared spectroscopy and boosting
  partial least squares.
\newblock Vibrational Spectroscopy \textbf{54}(1), 35--41 (2010)

\bibitem{43}
Trendafilov, N.T., Unkel, S., Krzanowski, W.: Exploratory factor and principal
  component analyses: some new aspects.
\newblock Kluwer Academic Publishers (2013)

\bibitem{2}
Wold, S., Esbensen, K., Geladi, P.: Principal component analysis.
\newblock Chemometrics and Intelligent Laboratory Systems \textbf{2}(1), 37--52
  (1987)

\bibitem{16}
Wold, S., Sjostrom, M., Eriksson, L.: Pls-regression: a basic tool of
  chemometrics.
\newblock Chemometrics and Intelligent Laboratory Systems \textbf{58}(2),
  109--130 (2001)

\bibitem{29}
Xu, L., Jiang, J.H., Zhou, Y.P., Wu, H.L., Shen, G.L., Yu, R.Q.: Mccv stacked
  regression for model combination and fast spectral interval selection in
  multivariate calibration.
\newblock Chemometrics and Intelligent Laboratory Systems \textbf{87}(2),
  226--230 (2007)

\bibitem{9}
Ye, J.: Least squares linear discriminant analysis.
\newblock In: Proceedings of the 24 International Conference on Machine
  Learning, pp. 1087--1093 (2007)

\bibitem{13}
Ye, J., R, J., Q, L.: Two-dimensional linear discriminant analysis.
\newblock Advances in Neural Information Processing Systems pp. 1431--1441
  (2005)

\bibitem{20}
Zhang, M.H., Xu, Q.S., Massart, D.L.: Boosting partial least squares.
\newblock Analytical Chemistry \textbf{77}(5), 1423--1431 (2005)

\bibitem{12}
Zheng, W., Zhao, L., Zou, C.: An efficient algorithm to solve the small sample
  size problem for lda.
\newblock Pattern Recognition \textbf{37}(5), 1077--1079 (2004)

\bibitem{32}
Zhou, Z.H., Wu, J., Tang, W.: Ensembling neural networks: Many could be better
  than all.
\newblock Artificial intelligence \textbf{137}(1), 239--263 (2002)

\end{thebibliography}

\end{document}